# Design of a Framework to Facilitate Decisions Using Information Fusion

**Tamer M. Abo Neama**

Military Technical College
*Tamer_abonema@yahoo.com*

**Ismail A. Ismail**

Arab Academy for Science, Technology and Maritime Transport
*Ismail.ghafar@aast.edu*

**Tarek S. Sobh**

Information Systems Department, Egyptian Armed Forces
*tarekvoice@yahoo.com*

**M. Zaki**

Azhar University
*azhar@enu.eg*

## Abstract

Information fusion is an advanced research area which can assist decision makers in enhancing their decisions. This paper aims at designing a new multi-layer framework that can support the process of performing decisions from the obtained beliefs using information fusion. Since it is not an easy task to cross the gap between computed beliefs of certain hypothesis and decisions, the proposed framework consists of the following layers in order to provide a suitable architecture (ordered bottom up):

1. A layer for combination of basic belief assignments using an information fusion approach. Such approach exploits Dezert-Smarandache Theory, DSmT, and proportional conflict redistribution to provide more realistic final beliefs.

2. A layer for computation of pignistic probability of the underlying propositions from the corresponding final beliefs.

3. A layer for performing probabilistic reasoning using a Bayesian network that can obtain the probable reason of a proposition from its pignistic probability.

4. Ranking the system decisions is ultimately used to support decision making.

    A case study has been accomplished at various operational conditions in order to prove the concept, in addition it pointed out that:

1. The use of DSmT for information fusion yields not only more realistic beliefs but also reliable pignistic probabilities for the underlying propositions.

2. Exploiting the pignistic probability for the integration of the information fusion with the Bayesian network provides probabilistic inference and enable decision making on the basis of both belief based probabilities for the underlying propositions and Bayesian based probabilities for the corresponding reasons.

A comparative study of the proposed framework with respect to other information fusion systems confirms its superiority to support decision making.

## Keywords

Information Fusion, Bayesian networks, Belief combination, Pignistic Probability, Decision Making.

## 1- Introduction

The fusion of information arises in many fields of applications nowadays (especially in defense, medicine, finance, geo-science, economy, etc). To deal with the challenges of such applications an Information Fusion and Probabilistic Decision Making Framework has been designed to cross the large gap between beliefs and decision making. At the bottom layer of such framework we use DSmT for information fusion that yields not only more realistic beliefs but also reliable pignistic probabilities for the underlying propositions. The second layer performs the pignistic probability, BetP{.}, computation. Such BetP{.} is used for the integration of the information fusion with the Bayesian network that occupies the third layer. Such architecture provides probabilistic inference and enables decision making on the basis of both belief based probabilities for the underlying propositions and Bayesian based probabilities for the corresponding reasons Figure (1).

Actually the proposed framework builds up smoothly a probability structure that can be passed to the higher layer. The event (observation) of highest probability is passed to the higher layer where a Bayesian network can provide probabilistic reasoning [Weise et al., 1993]. For convenience, this approach is used to perform reasoning rather than using descriptive logic or a truth maintenance technique. Finally the decisions at the human computer interface are ranked rather than being expressed flat in order to support the decision maker.

The paper is organized as follows: Section 2 explains the related work to the proposed framework, while section 3 presents the problem statement. Setion 4 expresses the architecture of the proposed framework. Section 5 presents the pignistic probability transformation. Section 6 discusses the probabilistic reasoning using Bayesian network. Section 7 discusses the results of an experimental case study. Finally, section 8 comprises the conclusion.

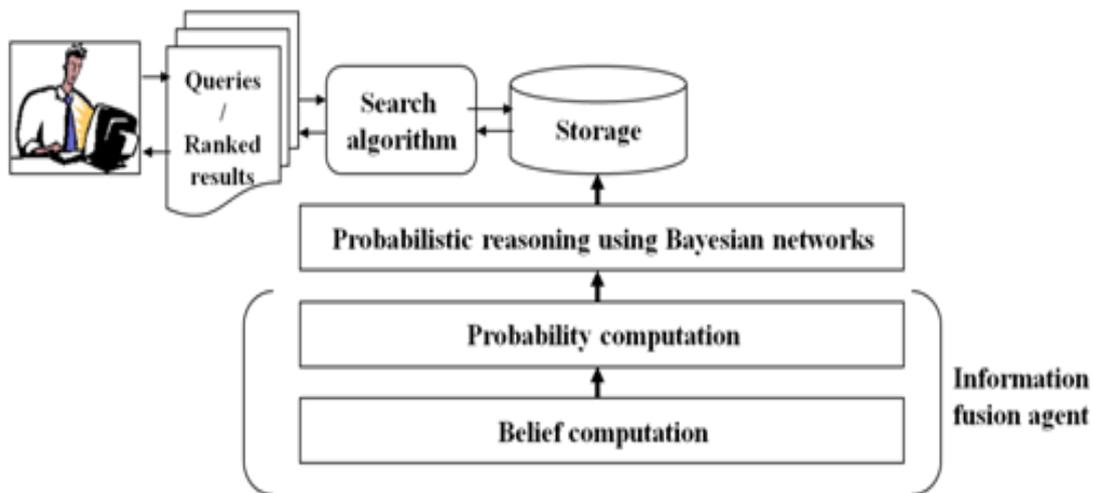

Figure (1) The proposed framework

## 2- Related Work

Information fusion (IF) is defined as the combination of data from disparate sources to produce an outcome that is superior to any provided by an individual source. An outcome typically includes an improvement in accuracy, higher confidence through complementary information, or improved performance in the presence of countermeasures [Blasch et al., 2002]. IF can occur on multiple levels [Klein, 2004]. Sensor-level fusion is the level at which relevant data is extracted from the source signal. Feature-level fusion is the combination of data to produce a composite feature vector that characterizes the object under test. Decision-level fusion is the layer that provides a projection of a future state of the object based on the feature vector provided, and the information presented to an operator to facilitate a human decision.

### 2.1 Military systems

The work of [Krenc et al., 2009] presents experiences related to a combination of two contrary approaches to information fusion: the first is typically deterministic ontology fusion and the second is based on the theory of evidence by Dezert and Smarandache (DSmT). It is the expectation of the authors that the appropriate synergy of these two approaches may bring satisfactory results when fusing diverse types of information originated from miscellaneous sensors. For this reason a concept of the combination of these two approaches has been presented and a comparison of hard-decision fusion, DSmT fusion and a combination of DSmT and ontology fusion algorithms has been established.

The authors of [Sumari et al., 2008] have designed and implemented a hierarchical multi-agent based information fusion system for decision making. The information fusion is implemented by applying a maximum score of the total sum of joint probabilities and is done by a collection of Information Fusion Agents (IFA) that forms a multiagent system. Information fusion products are displayed in graphical forms to provide comprehensive information regarding the military operation. By observing the graphics resulted from the information fusion, the commandant will have situational awareness and knowledge in order to make the most accurate strategic decision as fast as possible.

## 2.2 Decision support systems DSSs

The work of [Dezert et al., 2011] presented an extension of the multi-criteria decision making based on the Analytic Hierarchy Process (AHP) which incorporates uncertain knowledge for generating basic belief assignments (bba's). The combination of priority vectors corresponding to bba's related to each sub-criterion is performed using the Proportional Conflict Redistribution, PCR, which has been proposed in [Dezert et al., 2011] forplausible and paradoxical reasoning. The method presented, called DSmT-AHP, is illustrated on simple examples.

The multi-criteria decision-making (MCDM) problem concerns the elucidation of the level of preferences of decision alternatives through judgments made over a number of criteria [Beynon, 2005]. At the Decision-maker (DM) level, a useful method for solving MCDM problem must take into account opinions made under uncertainty and based on distinct criteria with different importance. Among the interesting solutions of MCDM problem there is the work made by [Beynon, 2005]. This work includes a method called DS/AHP which extended the Analytic Hierarchy Process (AHP) method of Saaty [Saaty, 1990] with Dempster- Shafer Theory (DST) of belief functions so that it can take into account uncertainty and to manage the conflicts between expert's opinions within a hierarchical model approach.

The authors of [Smarandache et al., 2009] have investigated the possibility to use (DSmT) of plausible and paradoxical reasoning for overcoming DST limitations. Their approach is referred to as DSmT-AHP method. In this case, DSmT allows managing efficiently the fusion of quantitative (or qualitative) uncertain and possibly highly conflicting sources of evidences and proposes new methods for belief computation.

## 2.3 Analysis and identification systems

The work of [Jousselme et al., 2003] has analyzed an identification algorithm in the evidence theory framework. The identification algorithm is composed of four main steps:

(1) Sensor reports are transformed into initial Basic Probability Assignments, BPA.

(2) The successive BPAs are combined through Dempster's rule.

(3) The resulting BPAs are approximated to avoid algorithm explosion.

(4) In parallel to step (3) a decision is taken on the identification/classification of an object from a database which is based on the maximum of pignistic probability criterion.

## 2.4 Discussion

The combination of statistical (probabilistic) reasoning and information fusion can afford powerful modeling and simulation tools that might be relied upon in various applications. Such applications may be military, political, analytical identification and decision support systems. Despite the fact that there is no formal architectural framework for modeling the underlying application there is a common agreement about an informal multi-layer architecture that consists of:

   a. The layer for calculating belief combinations and information fusion.
   b. The layer to find out the corresponding pignistic probability.
   c. The higher layer to perform probabilistic reasoning (in most of the cases using a variant of Bayesian networks).
   d. The top layer that can announce the undertaken decisions.

Actually, all the previous works have implemented only partially these layers either manually or semi automatically.

## 3- Problem statement

Decision-making from heterogeneous, poorly reliable and conflicting information is a major challenge in most areas ofscience and engineering. Nowadays, there is a possibility to access an increasing amount of information and in some cases to make use of all that information to be able to make an informed and successful decision. The goal of high-level information fusion is to provide effective decision-support regarding situations.

In this paper, this problem will be addressed by introducing a design of a framework to facilitate decisions from computed beliefs using information fusion and the theory of belief functions. The basic elements of the underlying framework are pointed out as follows:

1) **Given:** The basic beliefs of the underlying hypotheses.

2) **Use:** An information fusion agent to obtain hypothetical beliefs, m(.) from the available data using DSmT, taking into consideration the constraints that:

$$m(\emptyset) = 0, and \sum_{A \subseteq \Theta} m(A) = 1 \quad (1)$$

Then compute the pignistic probabilities BetP{.} from the obtained beliefs, m(X) using:

$$\text{BetP}\{A\} = \sum_{x \in D^\Theta} \frac{C_M(X \cap A)}{C_M(X)} m(X) \quad (2)$$

Where $A \in G^\Theta$; G is the space of beliefs and C denotes the cardinality. Performing probabilistic reasoning using a Bayesian network based on the values of that pignistic probability.

3) **Get:** Ranked decisions; $\{d_1, d_2 \ldots d_p \ldots d_q; 1 \leq p \leq q\}$, about the events that have been included in raw data in order to support the decision maker.

## 4- The architecture of the proposed framework

The proposed framework aims at providing a cause from the corresponding events (effects). To fulfill such aim it consists of several components that are integrated together, Figure (1). Figure (2) represents the transformation from beliefs to decisions. The relations between the proposed framework components are pointed out in the following:

1) *Information fusion using beliefs*

   In this module the belief functions are combined using DSmT Theory. Accordingly, a set of the required beliefs are obtained, Figure (2).

2) *Probability computation*

   It takes belief functions as input, makes necessary transformations and provides pignistic probabilities as its output.

3) *Bayesian network*

   To provide probabilistic reasoning by taking the probabilities of observations / (symptoms) in order to compute the probability of evidence / (disease) and the system decisions are ranked according to their weights.

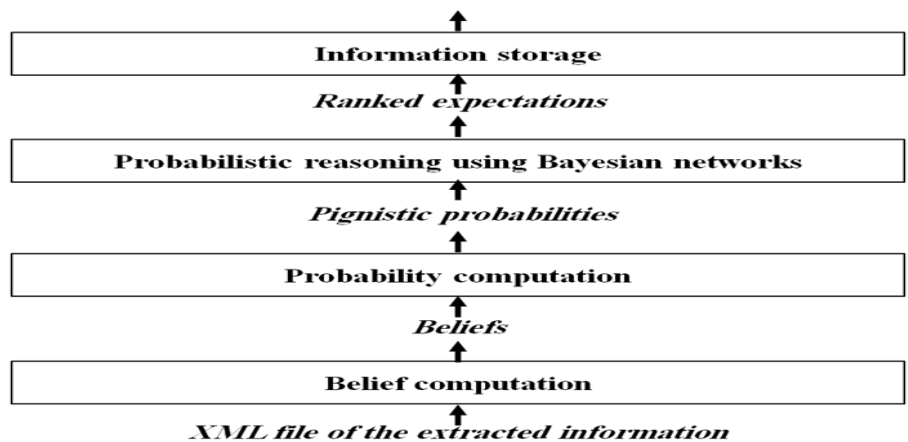

Figure (2) Process model of the proposed framework

## 5- Pignistic probability transformation

Based on [Dezert et al., 2004] in order to take a rational decision within the DSmT framework, it is then necessary to construct a pignistic probability function from any generalized basic belief assignment, m(.), drawn from the DSm rule of combination (the classic or hybrid rule). This generalized pignistic transformation (GPT) is defined by equation (2):

$$\forall A \in D^\Theta, \qquad \text{BetP}\{A\} = \sum_{x \in D^\Theta} \frac{C_M(X \cap A)}{C_M(X)} m(X)$$

Where $C_M(X)$ denotes the DSm cardinal of proposition X for the DSm model M of the problem under consideration. The decision about the solution of the problem is usually taken by the maximum of pignistic probability function BetP{.}.

It has been proven in that BetP{A} is a subjective probability measure satisfying the following axioms of the probability theory:

- **Axiom 1** (nonnegative): The (generalized pignistic) probability of any event A is bounded by 0 and 1, i.e. $0 \leqq P\{A\} \leqq 1$

- **Axiom 2** (unity): Any sure event (the sample space) has unity (generalized pignistic) probability, i.e. P{S} = 1

- **Axiom 3** (additively over mutually exclusive events): If A, B are disjoint (i.e. A∩B = ∅) then P(A ∪ B) = P(A) + P(B)

## 5.1 The DSm cardinality

One important notion involved in the definition of the generalized pignistic transformation (GPT) is the DSm cardinality [Dezert, 2003] [Dezert et al., 2004]. The DSm cardinality of any element $A \in D^\Theta$, denoted $C_M(A)$, corresponds to the number of parts of A in the Venn diagram of the problem (model $M$) taking into account the set of integrity constraints (if any), i.e. all the possible intersections due to the nature of the elements $\theta_i$. This intrinsic cardinality depends on the model $M$ (free, hybrid or Shafer's model). $M$ is the model that contains A, which depends both on the dimension $n = |\Theta|$ and on the number of parts of non-empty intersections present in its associated Venn diagram. One has $1 \leq C_M(A) \leq 2^n - 1$. $C_M(A)$ must not be confused with the classical cardinality |A| of a given set A (i.e. the number of its distinct elements) that is why a new notation is necessary here.

$C_M(A)$ is exactly equal to the sum of the elements of the row of $D_n$ corresponding to proposition A in the $u_n$ basis. Actually $C_M(A)$ is easy to compute by programming from the algorithm of generation of $D^\Theta$ [Dezert, 2003]. If one imposes a constraint that a set B from $D^\Theta$ is empty (i.e. we choose a hybrid model), then one suppresses the columns corresponding to the parts which compose B in the matrix $D_n$ and the row of B and the rows of all elements of $D^\Theta$ which are subsets of B, getting a new matrix $D'_n$ which represents a new hybrid model $M'$.

### *5.2 A 3D example with a given hybrid model*

[Dezert et al., 2004] said consider now a 3D example in which we force all possible conjunctions to be empty, but $\theta_1 \cap \theta_2$ according to the following Venn diagram shown in Figure (3).

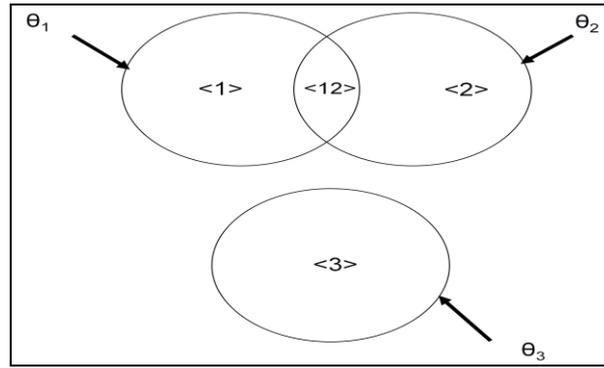

Figure (3) Venn diagram of a DSm hybrid model for a 3D frame [Dezert et al., 2004]

Then, one gets the following list of elements (with their DSm cardinal) for the restricted $D^\Theta$ taking into account the integrity constraints of this hybrid model:

Table (1) Cardinality $C_{M^f}(A)$ for the chosen hybrid model $M^f$ [Dezert et al., 2004]

| $A \in D^\Theta$ | $C_M(A)$ |
|---|---|
| $\alpha_0 \triangleq \emptyset$ | 0 |
| $\alpha_1 \triangleq \theta_1 \cap \theta_2$ | 1 |
| $\alpha_2 \triangleq \theta_3$ | 1 |
| $\alpha_3 \triangleq \theta_1$ | 2 |
| $\alpha_4 \triangleq \theta_2$ | 2 |
| $\alpha_5 \triangleq \theta_1 \cup \theta_2$ | 3 |
| $\alpha_6 \triangleq \theta_1 \cup \theta_3$ | 3 |
| $\alpha_7 \triangleq \theta_2 \cup \theta_3$ | 3 |
| $\alpha_8 \triangleq \theta_1 \cup \theta_2 \cup \theta_3$ | 4 |

## 6- Probabilistic reasoning using Bayesian network

Depending on the belief analysis, the pignistic probability of the corresponding event could be computed by the proposed model. However, in many situations such probability is not sufficient for the application user. For instance, if the application is medical it will represent the probability of a 'symptom', as advised by multiple experts but we still have the question: what is the probability of the 'disease', which caused such symptom?

The question is answered in the proposed framework model by making use of a Bayesian Network, BN, to represent the dependencies among variables in order to provide a concise specification of any joint probability distribution. The underlying BN takes the pignistic probability of an event as input and computes a corresponding expectation probability depending on both its topology and the conditional probability tables.

A BN as a directed graph, in which each node is annotated with quantitative probability information could perform the reasoning process. The full specification of BN is as follows:

1- A set of random variables makes up the nodes of the network. Variables may be discrete or continuous.

2- A set of directed links or arrows connects pairs of nodes. If there is an arrow from node X to node Y, X is said to be a parent of Y.

3- Each node X, has a conditional probability distribution P(X | *Parents*(X)) that quantities the effect of the parents on the node.

4- The graph has no directed cycles and hence is a directed, acyclic graph, or DAG).

The topology of the network specifies the conditional independence relationships that hold in the domain. The intuitive meaning of an arrow in a properly constructed network is usually that X has a direct influence on Y. It is usually easy for a domain expert to decide what direct influences exist in the domain- much easier, in fact, than actually specifying the probabilities themselves.

Once the topology of the Bayesian network is laid out, we need only to specify a conditional probability distribution for each variable, given its parents. Here the combination of the topology and the conditional distributions suffices to specify the full joint distribution for all the variables. Eventually, the output of BN is passed to the user interface in prder to support his current decisions.

## 7- Case study

## 7.1 Experimental setup

Figure (4) presents the environment (hardware, software, and the connection) of the implementation of the proposed framework model.

Its configuration can be described as follows:

1- *Hardware configuration*
   a. Server machine
      i. Intel XEON dual core processor
      ii. 4GB RAM
      iii. 2 x 750GB HD
      iv. Ethernet
   b. client machine
      i. Intel Core 2 Duo
      ii. 2 GB RAM
      iii. 80 GB HD
      iv. Ethernet
   c. Ethernet router
   d. Internet connection.

2- *Software configuration*
   a. Server machine
      i. OS: windows server for server machine
      ii. NetBeans IDE 7.0.
      iii. java development kit 1.6.
      iv. JADE agent software.
      v. GeNIe 2.0 (Bayesian network decision system)
   b. client machine
      i. OS : windows 7 for client machine.
      ii. java development kit 1.6.
      iii. JADE agent software.

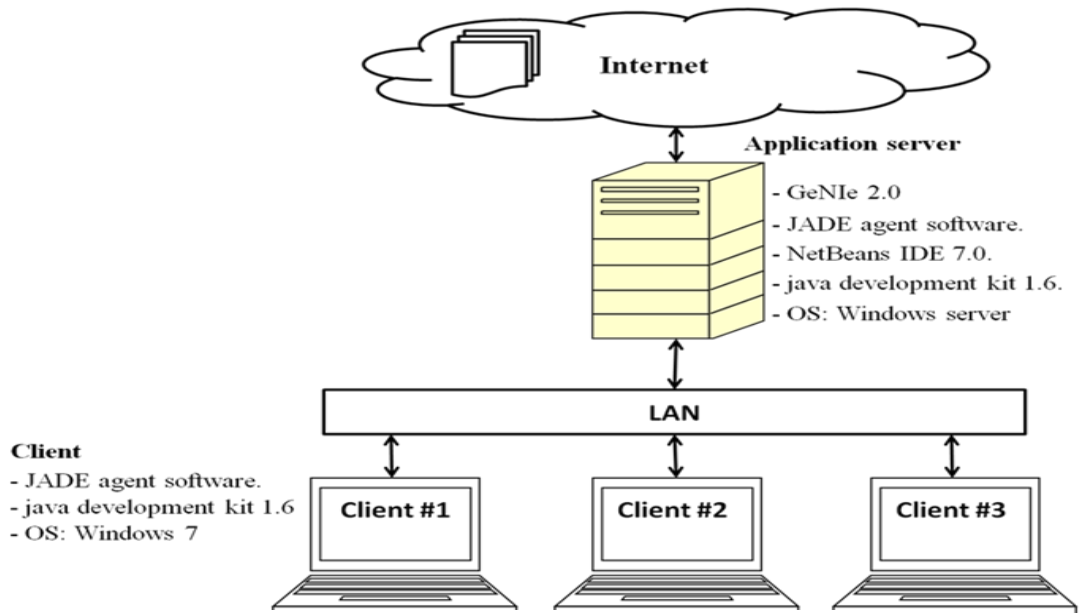

Figure (4) the proposed framework configuration

## 7.2 Case study details

The case study under consideration discusses the problem of international mediation between two countries to solve some regional dispute. Throughout the analysis of the raw data, the proposed framework has found out five countries that may perform concealed mediation to resolve that dispute.

To deal with this issue a divide and conquer approach has been exploited. Consequently, the hypotheses of the five countries are divided in two stages, the first stage takes into account three countries while the second stage comprises the three other countries, (one of the countries from the first stage along with the other two countries which have not been chosen before).

There are mainly two sources of information, namely, S1 and S2, where S1 represents Arabic sources (text documents) and S2 represents foreign sources.

The user in this case study is interested in:

1- Computing the probabilities of the basic belief assignments obtained from the raw data for each country participated in mediation.
2- Finding out the real motivation of the mediator.

The solution of this case study could be obtained using the following procedure:

### 7.2.1 The solution procedure

- *Basic belief assignment:* The mediation problem can be represented by the frame of discernment $\Theta = \{\theta_1, \theta_2, \theta_3, \theta_4, \theta_5\}$, where $\theta_1, \theta_2, \theta_3, \theta_4$, and $\theta_5$ are the hypotheses E, F, G, R, and U respectively. Since the data has the constraint hat both E and F are related and could cooperate with each otheras well as E and R are similarly related (intersected) then the DSm hybrid model, $M(\Theta)$ is applicable rather than Shafer's model. By considering the two information sources, S1 and S2 and applying the semantic network module on both of them, we could construct:

Table (2) Basic belief assignment (bba's) for stage 1

|  | E | F | G |
|---|---|---|---|
| *m*(S1) | 0.51 | 0.49 | 0.0 |
| *m*(S2) | **0.52** | **0.0** | **0.48** |

Where *m* (S$_i$), i=1, 2 represents the basic belief assignment (bba's), of the underlying information source.

- *Application of the hybrid model:* The choice of the DSm hybrid model allows some propositional intersections, while others are empty. Here E∩F≠∅, while E ∩ G = E ∩ U = F ∩ G = R ∩ U = ∅, as well as all unions are not included. Then we performed the classic DSm rule in order to obtain the classic beliefs $m_{DSmC}(S_1)$ and $m_{DSmC}(S_2)$, Table (3). Upon computing the values of belief tables, the following two conditions are satisfied:

  - m(∅) = 0.
  - and, $\sum_{A \in G^\theta} m_{DSmC}(A) = 1$ , where G$^\theta$ is the space of beliefs i.e. A∈ {E, F, G, E∩F, E∩G, F∩G, E∩F∩G}.

Table (3) Classic DSm beliefs in stage 1

| | E | F | G | E ∩ F | E ∩ G | F ∩ G | E ∩ F ∩ G |
|---|---|---|---|---|---|---|---|
| m$_{DSmC}$ | 0.265 | 0.0 | 0.0 | 0.255 | 0.245 | 0.235 | 0.000 |

Depending on the classic DSm beliefs, a proportional conflict redistribution is performed to yields m$_{PCR5}$ as illustrated in Table (4)

- *Proportional conflict redistribution:* To execute proportional conflict redistribution PCR, we transferred (at stage 1), $m_{DSmC}$(E∩G)=0.245 to E and G and $m_{DSmC}$(F∩G)=0.235, to F and G proportionally.

Table (4) Proportional conflict redistribution for stage 1

| | E | F | G | E ∩ F | E ∩ G | F ∩ G | E ∩ F ∩ G |
|---|---|---|---|---|---|---|---|
| m$_{PCR5}$ | 0.391 | 0.119 | 0.235 | 0.255 | 0.000 | 0.000 | 0.000 |

From Table (4) it is obvious that E has the highest m$_{PCR5}$ value. Therefore E is chosen and added to R and U to form the second stage that starts with bba's:

Table (5) Basic belief assignment (bba's) for stage 2

| | E | R | U |
|---|---|---|---|
| m(S1) | 0.5 | 0.5 | 0.0 |
| m(S2) | 0.48 | 0.0 | 0.52 |

Where m (Si), i=1, 2 represents the basic belief assignment (bba's), of the underlying information source. From the bba's, the classic DSm beliefs are computed and reported in Table (6):

Table (6) Classic DSm beliefs in stage 2

| | E | R | U | E ∩ R | E ∩ U | R ∩ U | E ∩ R ∩ U |
|---|---|---|---|---|---|---|---|
| m$_{DSmC}$ | 0.24 | 0.0 | 0.0 | 0.24 | 0.26 | 0.26 | 0.000 |

To carry out the PCR, we transferred (at stage 2), $m_{DSmC}(E \cap U)=0.26$ to E and U and $m_{DSmC}(R \cap U)=0.26$, to R and U, proportionally.

Table (7) Proportional conflict redistribution for stage 2

|  | E | R | U | E ∩ R | E ∩ U | R ∩ U | E ∩ R ∩ U |
|---|---|---|---|---|---|---|---|
| $m_{PCR5}$ | 0.368 | 0.128 | 0.266 | 0.240 | 0.000 | 0.000 | 0.000 |

The combination of stages (1) and (2) leads to Table (8) with final belief values. In that aggregated table the belief values, mPCR5 for E are ORed together, yielding a value of 0.391.

Table (8) The Combination of $m_{PCR5}$ for stages (1) and (2)

|  | E | F | G | R | U | E ∩ F | E ∩ R |
|---|---|---|---|---|---|---|---|
| $m_{PCR5}$ | 0.391 | 0.119 | 0.235 | 0.128 | 0.266 | 0.255 | 0.240 |

- *Pignistic probabilities:* The pignistic probability values for the underlying Θ are obtained from the available belief values, $m_{PCR5}$, by applying the following formula:

$$\forall A \in G^\Theta, BetP\{A\} = \sum_{x \in G^\Theta} \frac{C_M(X \cap A)}{C_M(X)} m(X)$$

Here $G^\Theta$ is the space of beliefs and $C_M(X)$ denotes the cardinality i.e. the number of the hypothesis parts in the Venn diagram of *M*. Therefore, the pignistic probability is calculated by substituting $C_M(X)$ with the corresponding value illustrated in the cardinality table which illustrated in Table (1), and similarly by substituting m(X) with its corresponding belief value from the proportional conflict redistribution belief value shown in Table (8).

Table (9) and Table (10) show the results of pignistic probabilities for the two stages. By combining the two tables one can obtain Table (11) in which BetP{E} is chosen by taking the higher corresponding value in the stage tables.

Table (9) Calculations of the pignistic probabilities for stage 1

|  | E | F | G | E ∩ F | E ∩ G | F ∩ G | E ∩ F ∩ G |
|---|---|---|---|---|---|---|---|
| BetP{.} | 0.708 | 0.571 | 0.236 | 0.511 | 0.000 | 0.721 | 0.000 |

Table (10) Calculations of the pignistic probabilities for stage 2

|  | E | R | U | E ∩ R | E ∩ U | R ∩ U | E ∩ R ∩ U |
|---|---|---|---|---|---|---|---|
| BetP{.} | 0.671 | 0.551 | 0.266 | 0.488 | 0.000 | 0.569 | 0.000 |

Table (11) The combination of pignistic probabilities for stages (1) and (2)

|  | E | F | G | R | U | E ∩ F | E ∩ R |
|---|---|---|---|---|---|---|---|
| BetP{.} | 0.708 | 0.571 | 0.236 | 0.552 | 0.266 | 0.511 | 0.488 |

- *Bayesian network:* The pair of expected "mediator" and its "associated probability" has been exploited as input to the BN, which can provide probabilistic reasoning that may help discovering the real motivation of a certain country to participate in the concluded mediation process. The BN consists of the probable mediator as an input node, the ability to take the decision to be a mediator as intermediate nodes and the motivating cause as the output node as shown in Figure (5). That figure indicates that, at the beginning both the political and the military motivations have equal probabilities (0.5) for both true and false causes.

However, when mediator E is introduced with its highest pignistic probability value, the corresponding causes are changed to be political motivation with true and false values (0.973, 0.027) while the military motivation is decayed to (0.08, 0.92) respectively.

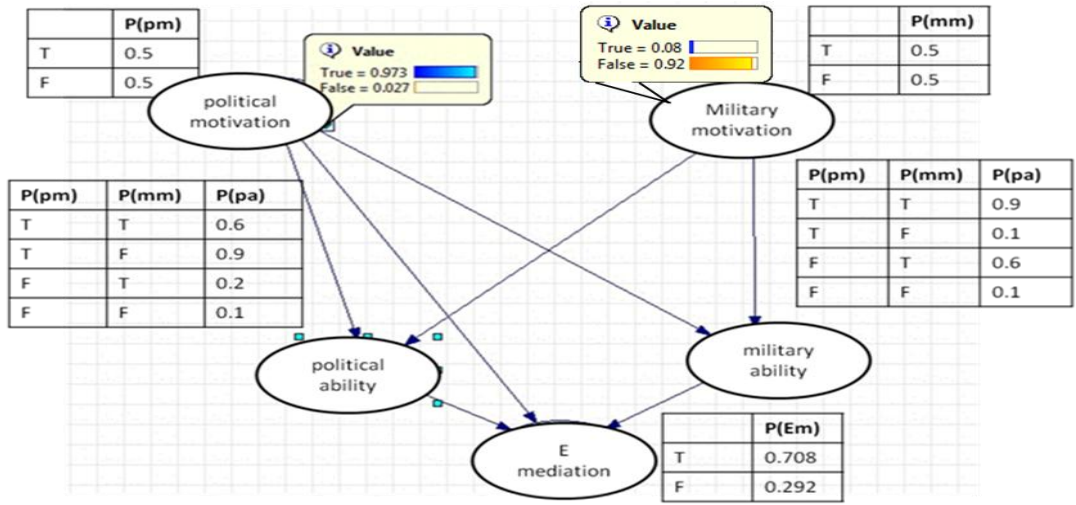

Figure (5) Bayesian Network for the case study

## 7.2.2 Discussion of the case study

By analyzing such case study (procedures and results) one can point out the following:

- General problems of information fusion in which Θ={θ₁,θ₂…θₙ}; where n ≥ 2, can be tackled by DSmT to compute the values of the required beliefs. If the problem dimensionality is high then can be divided into several subproblems. When the original problem has *n* dimensions, it can be divided into a number of three-dimensional problems so that $n \bmod 3 = r$; where r is the number of residual propositions. With this partitioning the data preprocessing becomes manageable and the size of the information fusion problem as well as its complexity is elegantly reduced.
- The problem partitioning should be carried out carefully, as follows:
  i. For n elements (propositions) get k three dimensional and r residues. Without loss of generality, in our case study k=1 and r=2 and it undergoes two independent stages (stage1 and stage 2).
  ii. In stage 1; the three dimensional fusion problem is solved and the belief results of that problem are given as:

$$b_{11}, b_{1j}...b_{1J}; \quad \sum_{j=1}^{J} b_{1j} = 1 \quad (3)$$

  with $b_{1j}$ is the final belief value of the jth hypothesis in the 1st stage, and J is the number of elements in the set of beliefs $m_{PCR5}$ for the first stage.
  iii. In stage 2 the hypothesis of the highest belief value is added to the two residues to form the second three dimensional problem. For that stage the belief results are:

$$b_{21}, b_{2q}...b_{2Q}; \quad \sum^{Q} b_{1q} = 1 \quad (4)$$

  With $b_{2q}$ is the belief value of the qth hypothesis in the 2nd stage and Q is the number of elements in the set of beliefs $m_{PCR5}$ for the second stage.
  By combining the values of the beliefs of both stages an aggregate belief table can be obtained.
  iv. BetP{.} is computed for stage 1

$$P_{11}, P_{12,}..., P_{1J} \quad (5)$$

  Similarly for stage 2 we can obtain:

$$P_{21}, P_{22},\ldots,P_{2Q} \quad (6)$$

By combing the probabilities in the two stages the final pignistic probabilities BetP{.} can be obtained as

$$P_1, P_2,\ldots, P_{(J+Q-1)} \quad (7)$$

In this case actually a common hypothesis exists with two different (conflicting) values. To resolve such conflict the BetP{.} for that hypothesis is chosen for the aggregated values by taking the higher pignistic probability in the stage tables.

## 7.3 Comparative study

A comparative study of the proposed framework with other information fusion systems is illustrated in Table (12). This comparison as such, has pointed out the significance of the proposed framework features. These features confirm the fact that the proposed framework only represents a complete realization for all the model layers, Figure (1). This ensures its superiority to support decision making. In this comparative study the proposed framework is compared with the identity fusion algorithm [Jousselme et al., 2003] and the multi-agent information fusion system [Sumari et al., 2008].

Table (12) Comparative study between the proposed framework and other information fusion systems

| Model / Property | the proposed framework | Identity fusion algorithm [Jousselme et al., 2003] | Multi-Agent Information Fusion Systems [Sumari et al., 2008] |
|---|---|---|---|
| **Main application area(s)** | - Anti-terrorism<br>- Spy war<br>- Security negotiations | Direct fleet support scenarios where raw data reports are time dependent | Military operations |
| **The information fusion technique** | Use of Dezert Smarandache Theory, DSmT.<br><br>Actually, DSmT is applicable for "both" free and hybrid models that permit $\theta_i$'s of $\Theta$ to be intersected. | Use of Dempster Shafer theory of evidence for combining information coming from different sources.<br><br>This theory is applicable "only" for free models in which $\theta_i$'s of $\Theta$ should be exclusive and exhaustive. | The theory of evidence is not taken into consideration; consequently, no beliefs are calculated to be relied upon. Information fusion is based on the JDL model that has been carried out in four levels. |
| **Information management to support decision making** | On the basis of two levels:<br><br>- Credal for combination of beliefs.<br>- Pignistic for supporting probabilistic reasoning.<br><br>The pignistic probability has been computed to specify the trust in belief functions after information fusion and to support the Bayesian network probabilistic reasoning. | Computes basic probability assignments, BPAs and the successive BPA's are combined through Dempster's rule. | This process is implemented by applying maximum score of the total sum of joint probabilistic fusion method. |

| **Decision making** | By performing probabilistic reasoning using a Bayesian network that can obtain the probable reason of a proposition from its pignistic probability. Thus the DSmT pignistic probability can serve two folds:<br>1-Accuracy measure for the underlying beliefs.<br>2-Input to the Bayesian network providing a conformal integration (i.e. decision might be changed when the basic beliefs are changed) between the belief and the probabilistic reasoning model. | Decision is taken for the identification/classification of an object from a database based on the maximum value of pignistic probability criterion. | Bayes formulation can produce an inference concerning an observed object viewed from existing events. Then the a posteriori conditional probability for each object calculated, one can decide the best estimated hypothesis by taking the greatest posteriori conditional probability value. |
|---|---|---|---|
| **Evidence based reasoning** | Can obtain the reason behind the believed hypothesis. | Cannot obtain the reason behind the believed hypothesis. | Does not depend on beliefs and does not use evidence theory. |
| **Ranking the output decisions** | Ranking the system decisions is ultimately used to support decision making | Decision is taken based on the maximum value of pignistic probability. | There is no ranking of output decisions; only one can decide the best estimated hypothesis by taking the greatest value of a posteriori conditional probability calculated by the Bayesian network. |

## 8- Conclusion

Since it is not an easy task to obtain high-level decisions from computed beliefs, the proposed framework has been designed and implemented as Information Fusion and Probabilistic Decision Making Framework. Such framework consistes of the following layers:

1. A layer that contains a belief computation agent for estimating the basic belief assignments.

2. A layer for combination of basic belief assignments using an information fusion approach. Such approach exploits Dezert-Smarandache Theory, DSmT, and proportional conflict redistribution to provide more realistic final beliefs.

3. A layer for computation of pignistic probability of the underlying propositions from the corresponding final beliefs.

4. A layer for performing probabilistic reasoning using a Bayesian network that can obtain the probable reason of a proposition from its pignistic probability.

5. A layer for ranking the system decisions is ultimately used to support decision making.

A case study is investigated in details. It has proved the concept of the proposed framework and indicated the following:

1. The use of DSmT for information fusion yields not only more realistic beliefs but also reliable pignistic probabilities for the underlying propositions.

2. Making use of pignistic probabilities to integrate information fusion and the Bayesian network. By this way we could provide probabilistic inference and enable decision making that exploits both belief based probabilities for the underlying propositions and conditional probabilities of BN for finding out the corresponding reasons.

3. A divide and conquer approach can considerably reduce the problem size. Such approach can transfer a problem with impractical large size due to large number of focal elements to a set of familiar 3D subproblems that can be easily tackled with reasonable time and space costs.

A comparative study with respect to other information fusion systems has pointed out the significance of the proposed framework which makes it able to outperform similar decision making systems.